\documentclass[conference]{IEEEtran}
\usepackage{times}

\usepackage[numbers]{natbib}
\usepackage{multicol}
\usepackage[bookmarks=true]{hyperref}
\usepackage{dblfloatfix}
\usepackage{amsmath,amsfonts,bm}

\def\eqref#1{equation~\ref{#1}}

\def\1{\bm{1}}

\DeclareMathAlphabet{\mathsfit}{\encodingdefault}{\sfdefault}{m}{sl}
\SetMathAlphabet{\mathsfit}{bold}{\encodingdefault}{\sfdefault}{bx}{n}

\usepackage{subcaption} %
\usepackage[skip=4pt,font=small,labelfont=bf]{caption}
\usepackage{graphicx}
\usepackage{amsmath}
\usepackage{algpseudocode}
\usepackage{algorithm}
\usepackage[algo2e]{algorithm2e}
\usepackage{multirow}
\usepackage[normalem]{ulem}
\usepackage{hyperref}
\usepackage{booktabs}
\usepackage{verbatim}

\newcommand{\MODEL}[0]{CLIP-Fields}

\newcommand{\xxnote}[3]{}
\ifx\hidenotes\undefined
  \renewcommand{\xxnote}[3]{\color{#2}{#1: #3}}
\fi

\hypersetup{
    colorlinks=true,
    linkcolor=blue,
    filecolor=magenta,      
    urlcolor=blue,
}

\newcommand\blfootnote[1]{%
  \begingroup
  \renewcommand\thefootnote{}\footnote{#1}%
  \addtocounter{footnote}{-1}%
  \endgroup
}

\title{\LARGE \bf

CLIP-Fields: Weakly Supervised Semantic Fields\\ for Robotic Memory
}

\author{
Nur Muhammad (Mahi) Shafiullah$^\dagger$$^1$\\
\and
Chris Paxton$^2$\thanks{$2.$ FAIR Labs}\\
\and
Lerrel Pinto$^1$\\
\and
Soumith Chintala$^2$ \\
\and
Arthur Szlam$^2$ \thanks{$\dagger$ Corresponding author, email: \texttt{mahi@cs.nyu.edu}}\\
}
\begin{document}
\maketitle
\thispagestyle{empty}
\pagestyle{empty}
\blfootnote{{$\dagger$ Corresponding author, email \href{mahi@cs.nyu.edu}{mahi@cs.nyu.edu}}}
\blfootnote{1. New York University \; 2. FAIR Labs}
\begin{abstract}
We propose \MODEL{}, an implicit scene model that can be used for a variety of tasks, such as %
segmentation, instance identification, semantic search over space, and view localization.  \MODEL{} 
learns a mapping from spatial locations to semantic embedding vectors.
Importantly, we show that this mapping can be trained with supervision coming only from web-image and web-text trained models such as CLIP, Detic, and Sentence-BERT; and thus uses no direct human supervision.
When compared to baselines like Mask-RCNN, our method outperforms on few-shot instance identification or semantic segmentation on the HM3D dataset with only a fraction of the examples.
Finally, we show that using \MODEL{} as a scene memory, robots can perform semantic navigation in real-world environments.
Our code and demonstration videos are available here: 
\url{https://mahis.life/clip-fields}
\end{abstract}
\section{Introduction}

In order to perform a variety of complex tasks in human environments, robots often rely on a spatial semantic memory~\cite{blukis2022persistent,min2021film,gervet2022navigating}.
Ideally, this spatial memory should not be restricted to particular labels or semantic concepts, would not rely on human annotation for each scene,
and would be easily learnable from commodity sensors like RGB-D cameras and IMUs.
However, existing representations are coarse, often relying on a preset list of classes and capturing minimal semantics~\cite{blukis2022persistent,gervet2022navigating}.
As a solution, we propose \MODEL{}, which builds an implicit spatial semantic memory using web-scale pretrained models as weak supervision.

Recently, representations of 3D scenes via neural implicit mappings have become practical ~\cite{sucar2021imap,sitzmann2019scene}.  Neural Radiance Fields (NeRFs)~\cite{mildenhall2021nerf}, and implicit neural representations more generally~\cite{ortiz2022isdf} can serve as differentiable databases of spatio-temporal information that can be used by robots for scene understanding, SLAM, and planning~\cite{li20223d, simeonov2022neural, chen2022neural, driess2022learning, ortiz2022isdf}.

Concurrently, web-scale weakly-supervised vision-language models like CLIP~\cite{radford2021learning} have shown  that the ability to capture powerful semantic abstractions from individual 2D images.
These have proven useful for a range of robotics applications, including object understanding~\cite{thomason2022language} and multi-task learning from demonstration~\cite{shridhar2022cliport}. Their applications have been limited, however, by the fact that these trained representations assume a single 2D image as input; it is an open question how to use these together with 3D reasoning. %

In this work, we introduce a method for building weakly supervised semantic neural fields, called \MODEL{}, which combines the advantages of both of these lines of work. \MODEL{} is intended to serve as a queryable 3D scene representation, capable of acting as a spatial-semantic memory for a mobile robot. We show that \MODEL{} is capable of open-vocabulary segmentation and object navigation in a 3D scene using only pretrained models as supervision.

\begin{figure}[tb!]
\includegraphics[width=\columnwidth]{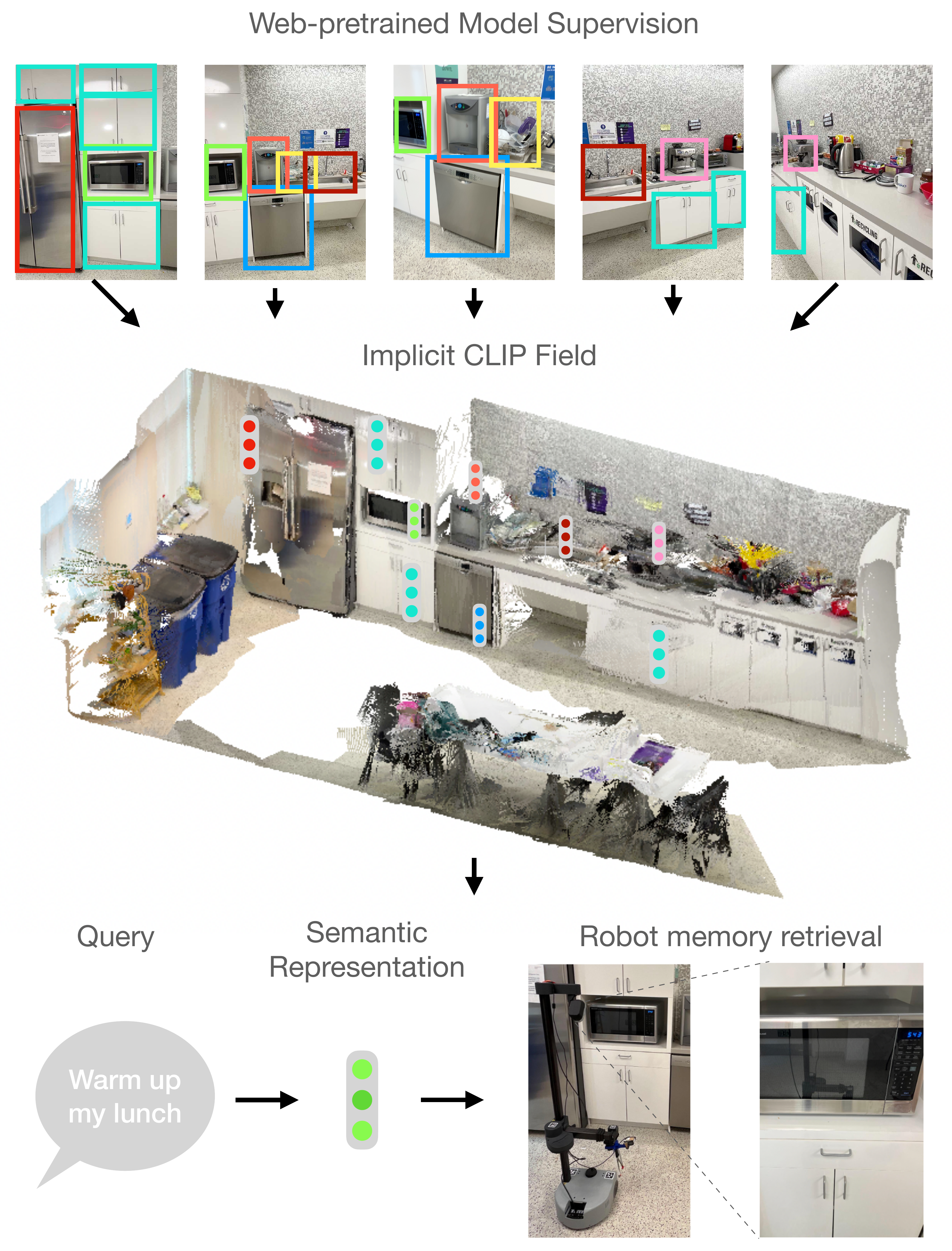}
\caption{Our approach, \MODEL{}, integrates multiple views of a scene and can capture 3D semantics from relatively few examples. This results in a scalable 3D semantic representation that can be used to infer information about the world from relatively few examples and functions as a 3D spatial memory for a mobile robot.
}
\label{fig:cover}
\vskip -0.5cm
\end{figure}

Our key idea is to build a mapping from locations in space $g(x, y, z): \mathbb{R}^3 \rightarrow \mathbb{R}^d$ that serves as a generic differentiable spatial database. This dataset is trained to predict features from a set of off-the-shelf vision-language models trained on web-scale data, which give us weak supervision. 
This map is trained on RGB-D data using a contrastive loss which encourages similarity between features predicted at specific spatial locations.

Thus, from the point of view of a robot using \MODEL{} as a spatial database for scene-understanding, training $g$ itself can be entirely self-supervised: the full pipeline, including training the underlying image models, need not use any explicit supervision.  On the other hand, as we show in our experiments, even without any explicit supervision, the spatial database $g$ can naturally capture scene-specific information. %

We demonstrate our method  on tasks such as instance segmentation and identification.
Furthermore, we give qualitative examples of image-view localization, where we need to find the spatial coordinates corresponding to an image and localizing text descriptions in space. 
Finally, we demonstrate \MODEL{} on a real robot by having the robot move to look at various objects in 3D given natural language commands.
These experiments show how \MODEL{} could be used to power a range of real-world applications by capturing rich 3D semantic information in an accessible way. %

\section{Related work}

\textbf{Vision-Language Navigation.} Much recent progress on vision-language navigation problems such as ALFRED~\cite{shridhar2020alfred} or RXR~\cite{ku-etal-2020-room} has used spatial representations or structured memory as a key component to solving the problem~\cite{min2021film,blukis2022persistent,wang2021structured,gadre2022continuous}.
HLSM~\cite{blukis2022persistent} and FiLM~\cite{min2021film} are built as the agent moves through the environment, and rely on a fixed set of classes and a discretization of the world that is inherently limiting. 
By contrast, \MODEL{} creates an embedding-dependant implicit representation of a scene, removing dependency on a fixed set of labels and hyperparameters related to environment discretization.
Other representations~\cite{wang2021structured} do not allow for 3D spatial queries, or rely on dense annotations, or accurate object detection and segmentation~\cite{gadre2022continuous,chen2020scanrefer,azuma_2022_CVPR}.

Concurrently with our work, NLMap-SayCan~\cite{chen2022open} and VLMaps~\cite{huang2022visual} proposed two approaches for real-world vision-language navigation. 
NLMap-SayCan uses a 2D grid-based map and a discrete set of objects predicted by a region-proposal network~\cite{chen2022open}, while \MODEL{} can make predictions at different granularities. 
VLMaps~\cite{huang2022visual} use a 2D grid-based representation and operate on a specific, pre-selected set of object classes. 
By contrast, \MODEL{} can operate on 3D data, allowing the agent to look up or down to find objects.
All three methods assume the environment has been explored, but both~\cite{chen2022open} and~\cite{huang2022visual} look at predicting action sequences, while we focus on the problem of building an open-vocabulary, queryable 3D scene representation.

\textbf{Pretrained Representations.} Effective use of pretrained representations like CLIP~\cite{radford2021learning} seems crucial to deploying robots with semantic knowledge in the real world.
Recent works have shown that it is possible to use supervised web image data for self-supervised learning of spatial representations. 
Our work is closely related to \cite{chaplot2021seal}, where the authors show that a web-trained detection model, along with spatial consistency heuristics, can be used to annotate a 3D voxel map.
That voxel map can then be used to propagate labels from one image to another.
Other works, for example \cite{datta2022episodic}, use models specifically trained on indoor semantic segmentation to build semantic scene data-structures.

Cohen et al.~\cite{cohen2022my} looks at personalizing CLIP for specific users and rare queries, but does not build 3D spatial representations conducive to robotics applications, and instead functions on the level of individual images.

\textbf{Implicit Representations.}
There is a recent trend towards using NeRF-inspired representations as the spatial knowledge base for robotic manipulation problems~\cite{simeonov2022neural,driess2022learning}, {but so far this has not been applied to open-vocabulary object search.}
As in \cite{zhi2021ilabel, sucar2021imap, vora2021nesf, kobayashi2022decomposing, tschernezki2022neural}, we use a mapping (parameterized by a neural network) that associates to an $(x, y, z)$ point in space a vector with semantic information.
In those works, the labels are given as explicit (but perhaps sparse) human annotation, whereas, in this work, the annotation for the semantic vector are derived from weakly-supervised web image data.

\textbf{Language-based Robotics.} Several works \cite{shridhar2022cliport, thomason2022language} have shown how features from weakly-supervised web-image trained models like CLIP~\cite{radford2021learning} can be used for robotic scene understanding.  Most closely related to this work is \cite{ha2022semantic}, which uses CLIP embeddings to label points in a single-view 3D space via back-projection. 
In that work, text descriptions are associated with locations in space in a two step process. In the first step, using an ViT-CLIP attention-based relevancy extractor, a given text description is localized in a region on an image; and that region is back-projected to locations in space (via depth information). 
In the second step, a separately trained model decoupled from the semantics converts the back-projected points into an occupancy map.
In contrast, in our work, CLIP embeddings are used to directly train an implicit map that outputs a semantic vector corresponding to each point in space. 
One notable consequence is that our approach integrates semantic information from multiple views into the spatial memory; for example in Figure \ref{fig:zero_shot_semseg} we see that more views of the scene lead to better zero-shot detections.

\section{Background}
\label{sec:background}
In this section, we provide descriptions of the recent advances in machine learning that makes \MODEL{} possible.

\paragraph{Contrastive Image-Language Pretraining} This pretraining method, colloquially known as CLIP \cite{radford2021learning}, is based on training a pair of image and language embedding networks such that an image and text strings describing that image have similar embeddings.
The CLIP model in \cite{radford2021learning} is trained with a large corpus of paired image and text captions with a contrastive loss objective predicting which caption goes with which image.
The resultant pair of models are able to embed images and texts into the same latent space with a meaningful cosine similarity metric between the embeddings.
We use CLIP models and embeddings heavily in this work because they can work as a shared representation between an object's visual features and its possible language labels.

\paragraph{Open-label Object Detection and Image Segmentation}
Traditionally, the objective of object detection and semantic segmentation tasks has been to assign a label to each detected object or pixels.
Generally, these labels are chosen out of a set of predefined labels fixed during training or fine-tuning.
Recently, the advent of open-label models have taken this task to a step further by allowing the user to define the set of labels during run-time with no extra training or fine-tuning.
Such models instead generally predict a CLIP embedding for each detected object or pixel, which is then compared against the label-embeddings to assign labels.
In our work, we use Detic~\cite{zhou2022detecting} pretrained on ImageNet-20k as our open-label object detector.
We take advantage of the fact that besides the proposed labels, Detic also reports the CLIP image embedding for each proposed region in the image.

\paragraph{Sentence Embedding Networks for Text Similarity}
CLIP models are pretrained with image-text pairs, but not with image-image or text-text pairs.
As a result, sometimes CLIP embeddings can be ambiguous when comparing similarities between two images or pieces of texts.
To improve \MODEL{}' performance on language queries, we also utilize language model pretrained for semantic-similarity tasks such as Sentence-BERT~\cite{reimers-2019-sentence-bert}.
Such models are pretrained on a large number of question-answer datasets.
Thus, they are also good candidates for generating embeddings that are relevant to answering imperative queries.

\paragraph{Neural Fields} Generally, Neural Fields refer to a class of methods using coordinate based neural networks which parametrize physical properties of scenes or objects across space and time~\cite{xie2022neural}.
Namely, they build a map from space (and potentially time) coordinates to some physical properties, such as RGB color and density in the case of neural radiance fields~\cite{mildenhall2021nerf}, or a signed distance in the case of instant signed distance fields~\cite{ortiz2022isdf}.
While there are many popular architectures for learning a neural field, in this paper we used Instant-NGP~\cite{mueller2022instant} as in preliminary experiments we found it to be an order of magnitude faster than the original architecture in \cite{mildenhall2021nerf}.

Note that a major focus of our work is using models pretrained on large datasets as-is -- to make sure \MODEL{} can take advantage of the latest advances in the diverse fields it draws from.
At the same time, while in our setup we haven't found a need to fine-tune any of the pretrained models mentioned here, we do not believe there is any barrier to do so if such is necessary.

\section{Approach}
In this section, we describe our concrete problem statement, the components of our semantic scene model, and how those components connect with each other.

\subsection{Problem Statement}
\label{sec:problem_statement}
We aim to build a system that can connect points of a 3D scene with their visual and semantic meaning.
Concretely, we design \MODEL{} to provide an interface with a pair of scene-dependent implicit functions $f, h: \mathbb R^3 \rightarrow \mathbb R^n$ such that for the coordinates of any point $P$ in our scene, $f(P)$ is a vector representing its semantic features, and $h(P)$ is another vector representing its visual features.
For ease of decoding, we constrain the output spaces of $f, h$ to match the embedding space of pre-trained language and vision-language models, respectively.
For the rest of this paper, we refer to such functions as ``spatial memory'' or ``geometric database'' since they connect the scene coordinates with scene information.

Given such a pair of functions, we can solve multiple downstream problems in the following way:
\begin{itemize}
    \item \textbf{Segmentation:} For a pixel in a scene, find the corresponding point $P_i$ in space. Use the alignment between a label embedding and $f(P_i)$ to find the label with the highest probability for that pixel.
    Segment a scene image by doing so for each pixel.
    \item \textbf{Object navigation:} For a given semantic query $q_s$ (or a visual query $q_v$) find the associated embeddings from our pretrained models, $e_s$ (respectively, $e_v$), and find the point in space that maximizes $e_s \cdot f(P^*)$ (or $e_v \cdot h(P^*)$). Navigate to $P^*$ using classic navigation stack.
    \item \textbf{View localization:} Given a view $v$ from the scene, find the image embedding $e_v$ of $v$ using the same vision-language model. Find the set of points with highest alignment $e_v\cdot h(P)$ in the scene.
\end{itemize}

While such a pair of scene-dependent functions $f, h$ would be straightforward to construct if we were given a dataset $\{(P, f(P), h(P) \mid P \in \text{scene}\}$, to make it broadly applicable, we create \MODEL{} to be able to construct $f, h$ from easily collectable RGB-D videos and odometry data.

\subsection{Dataset Creation}
\label{sec:dataset}
\begin{figure}[hbt!]
\vskip -0.25cm
\includegraphics[width=\columnwidth]{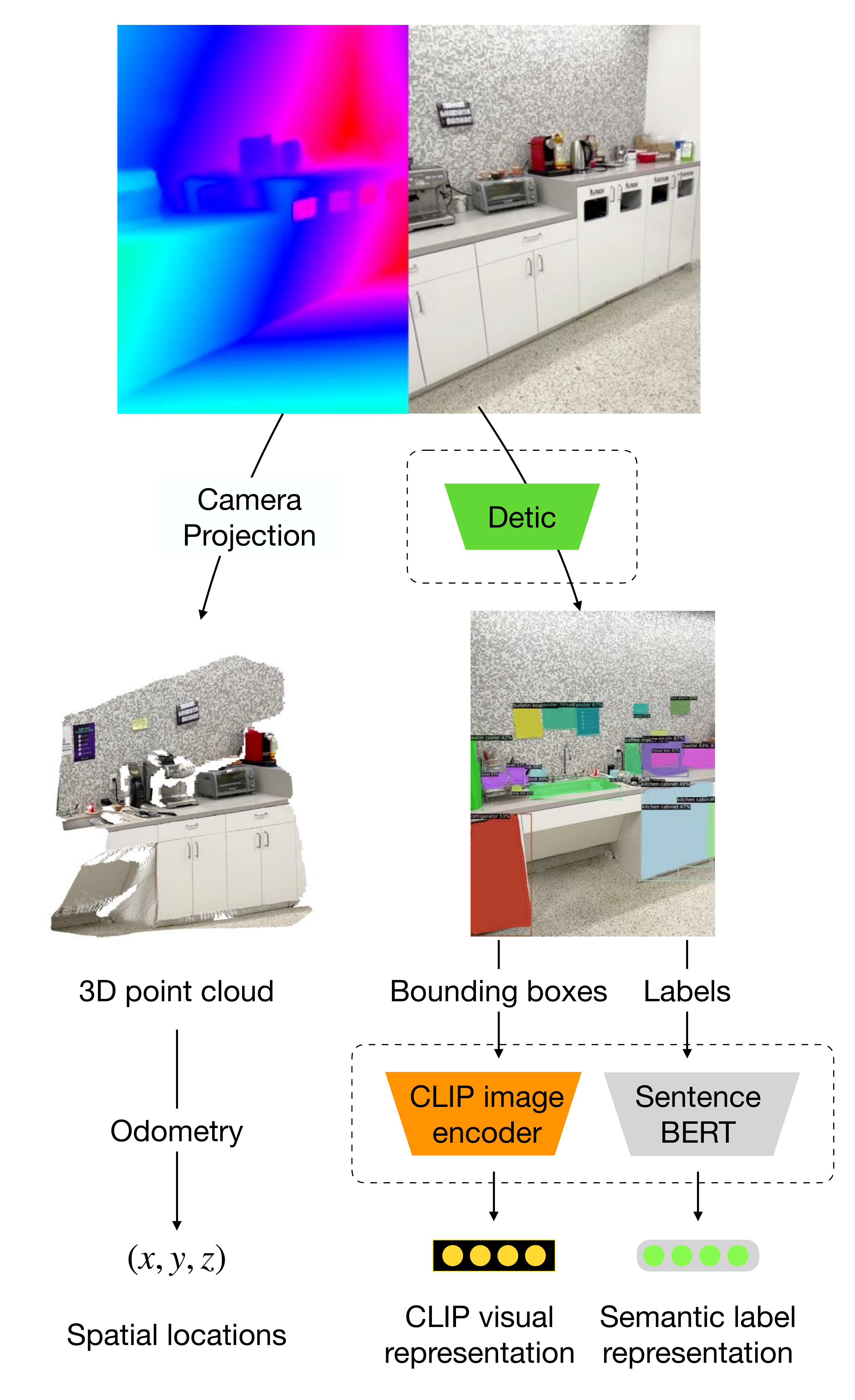}
\caption{Dataset creation process for \MODEL{} by processing each frame of a collected RGB-D video. Models highlighted by dashed lines are off-the-shelf pre-trained models, showing that we can train a real world \MODEL{} using no direct human supervision beyond pre-trained open label object detectors, large language models (LLMs) and visual language models (VLMs).
}
\label{fig:dataset}
\vskip -0.5cm
\end{figure}

We assume that we have a series of RGB-D images of a scene alongside odometry information, i.e. the approximate 6D camera poses while capturing the images.
As described in~\ref{sec:robot-exp}, we captured such a dataset using accessible consumer devices such as an iPhone Pro or iPads.
To train our model, we first preprocess this set of RGB-D frames into a scene dataset (Fig.~\ref{fig:dataset}).
We convert each of our depth images to pointclouds in world coordinates using the camera's intrinsic and extrinsic matrices.
Next, we label each of the points $P$ in the pointcloud with their possible representation vectors, $f(P), h(P)$.
When no human annotations are available, we use web-image trained object detection models on our RGB images.
We choose Detic~\cite{zhou2022detecting} as our detection model since it can perform object detection with an open label set.
However, this model can freely be swapped out for any other pretrained detection or segmentation model.
When available, we can also use human annotations for semantic or instance segmentations.

In both cases, we derive a set of detected objects with language labels in the image, along with their label masks and confidence scores.
We back-project the pixels included in the the label mask to the world coordinates using our point cloud.
We label each back-projected point in the world with the associated language label and label confidence score.
Additionally, we label each back-projected point with the CLIP embedding of the view it was back-projected from as well as the distance between camera and the point in that particular point.
Note that each point can appear multiple times in the dataset from different training images.

Thereby, we get a dataset with two sets of labels from our collected RGB-D frames and odometry information.
One set of label captures primarily semantic information, $D_{\text{label}} = \{(P, \text{label}_P, \text{conf}_P)\}$ where $\text{label}_P$ and $\text{conf}_P$ are just detector-given label and the confidence score to such label for each point.
The second set of labels captures primarily visual information, $D_{\text{image}} = \{(P, \text{clip}_P, \text{dist}_P)\}$, where $\text{clip}_P$ is the CLIP embedding of the image point $P$ was back-projected from, and $\text{dist}_P$ is the distance between $P$ and the camera in that image.
We then train \MODEL{} to efficiently combine the representations, encoding the points' semantic and visual properties in $g$.

\subsection{Model Architecture}
\label{sec:arch}
\begin{figure}[tb!]
    \centering
    \includegraphics[width=\linewidth]{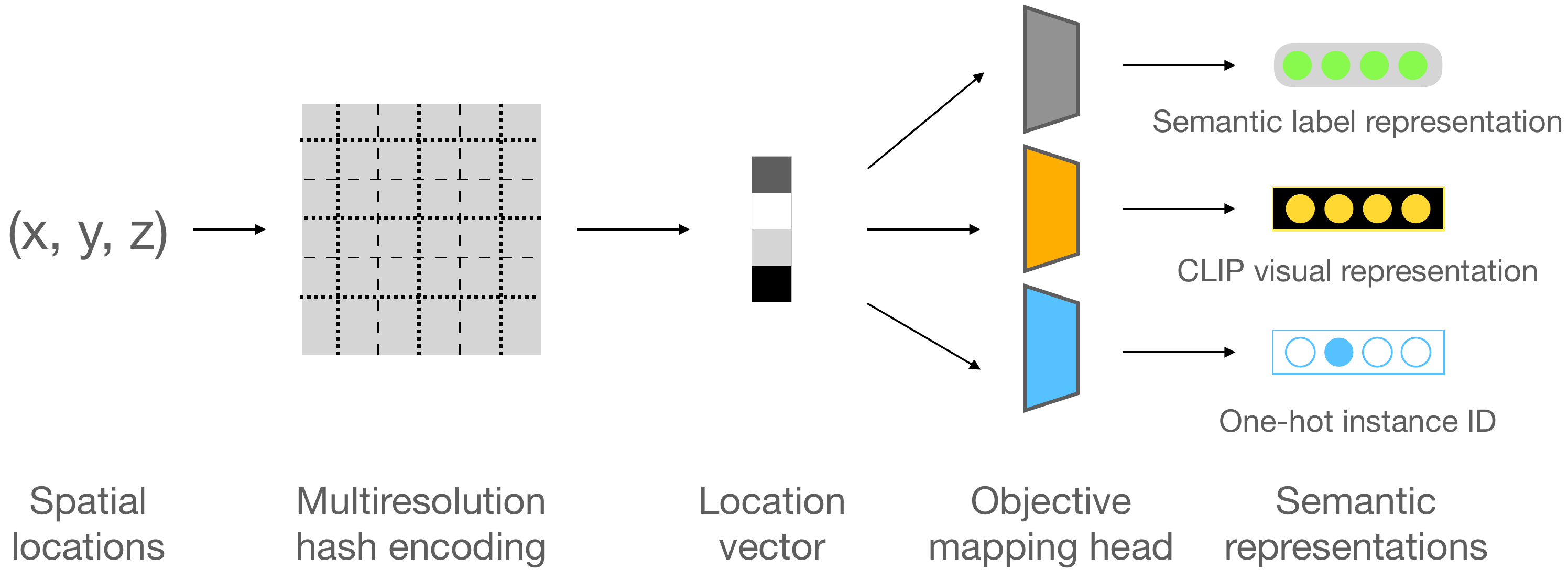}
    \caption{Model architecture for \MODEL{}. We use a Multi-resolution Hash Encoder~\cite{mueller2022instant} to learn a low level spatial representation mapping $\mathbb{R}^3 \rightarrow \mathbb{R}^{d}$, which is then mapped to higher dimensions and trained with contrastive objectives.}
    \label{fig:architecture}
    \vskip -0.5cm
\end{figure}

\MODEL{} can be divided into two components: a trunk $g: \mathbb{R}^3\rightarrow \mathbb{R}^d$, which maps each location $(x, y, z)$ to a representation vector, and individual heads, one for each one of our objectives, like language or visual representation retrieval.  See Figure \ref{fig:architecture} for an overview.

We can parameterize $g$ with any neural field architecture; in \MODEL{} we use multi-resolution hash encoding (MHE) as introduced in Instant-NGP~\cite{mueller2022instant}, with $d=144$.
MHEs build an implicit representation over coordinates with a feature-pyramid like structure, which can flexibly maintain both local and global information, unlike purely voxel-based encodings which focuses on local structures only.
We primarily use the MHE over other implicit field representations because we found that they train significantly faster in our datasets.
The objective-specific heads are simple two-layer MLPs with ReLU nonlinearities that map the $144$ dimensional outputs of $g$ into higher dimensions which depend on the associated objective. These include $\text{head}_s$ that outputs a vector that matches a natural language description of what is at the point in space, 
and $\text{head}_v$ that matches the visual appearance of the object occupying that point in space.
Optionally, we can include an instance identification head if we have the appropriate labels to train it.

\subsection{Objectives}
\label{sec:objectives}
The functions $f, h$ in our implicit scene model can be simultaneously trained with multiple objectives.
Each objective trains an implicit function that maps from real world locations in $\mathbb R^3$ to the objective space.
\MODEL{} are trained on a specific scene with a contrastive loss, similar to CLIP~\cite{radford2021learning}.
While training the contrastive loss objective, we also take into consideration the associated label weights.
For the contrastive loss calculation, the loss is weighted by the label confidence (for semantic labels, like label embeddings from SentenceBERT~\cite{reimers-2019-sentence-bert}), or negative exponential of distance from camera to point (for visual labels from CLIP~\cite{radford2021learning} embeddings).
Additionally, as is standard practice, we scale the dot product of the predicted and the ground truth embeddings by a learned temperature value.
We use the following training objectives:

\noindent\textbf{Semantic Label Embedding:} This objective trains the function encoding the semantic information of a 3D point as a $n$-dimensional representation vector.
We train this using the assigned natural language labels to each point.
We first convert each label to a semantic vector using a pre-trained language model trained to compare semantic similarity, such as CLIP~\cite{radford2021learning} or Sentence-BERT~\cite{reimers-2019-sentence-bert}.
In this paper we used Sentence-BERT for these language features with $n = 768$.

Mathematically, let us assume that $P$ is the point where we are calculating the loss, $P^-$ are points with a different semantic label, $f = \text{head}_s \circ g$ is the associated semantic encoding function, $\mathcal F$ is a pre-trained semantic language encoder, $c$ is the confidence associated with the label at $P$, and $\tau$ is a temperature term, then the semantic label loss is:
\begin{equation*}
\text{L}_{\mathcal{L}}(P, f(P)) = -\text{c}\log \frac{\exp{\left( f(P)^T \mathcal{F}(\text{label}_P) /\tau \right )} }{\sum_{P^-}\exp{\left( f(P)^T \mathcal{F}(\text{label}_{P^-}) /\tau \right )}}   
\end{equation*}

\noindent\textbf{Visual Feature Embedding:} This objective trains the embedding of the language-aligned visual context of each scene point into a single vector, akin to CLIP~\cite{radford2021learning}.
We define the visual context of each point as a composite of the CLIP embedding of each RGB frame this point was included in, weighted by the distance from camera to the point in that frame.
If it is possible to do so from the given annotation, we limit the image embedding to only encode what is in the associated object's bounding box.
Detic~\cite{zhou2022detecting}, for example, produces embeddings for region proposals for each detected objects, which we use.
In this paper's experiments, we use the CLIP ViT-B/32 model embeddings, giving the visual features $512$ dimensions.

Similar to the previous objective, given CLIP visual embedding $C$s associated with the points, the mapping $h = \text{head}_v \circ g$, the distance between camera and the positive point $d_P$, and temperature term $\tau$, the visual context loss $\text{L}_{\text{C}}$, is:
\begin{equation*}
\text{L}_{\text{C}}(P, h(P)) = -e^{-d_P}\log  \frac{\exp{\left( h(P)^T \text{C}_P /\tau \right )} }{\sum_{P^-}\exp{\left( h(P)^T \text{C}_{P^-} /\tau \right )}},
\end{equation*}

\noindent\textbf{Auxilary objectives like Instance Identification}: This optional head projects the point representation to a one-hot vector identifying its instance.
We use this projection head only in the cases where we have human labeled instance identification data from the scene, and the projection dimension is number of identified instances, plus one for unidentified instances.
Instance identification one-hot vectors are trained with a simple cross-entropy loss $L_I$.

Then, the final loss for \MODEL{} becomes
\[L = \text{L}_{\mathcal{L}} + \text{L}_{\text{C}} + \alpha L_I\]
where $\alpha$ is a normalizing hyper-parameter to bring the cross-entropy loss to a comparable scale of the contrastive losses.
\subsection{Training}
Our models are trained with the datasets described in Sec.~\ref{sec:dataset}.
We train the implicit maps simultaneously with the contrastive losses described in Sec.~\ref{sec:objectives}.
Under this loss, each embedding is pushed closer to positive labels and further away from negative labels.
For the label embedding head, the positive example is the semantic embedding of the label associated with that point, while negative examples are semantic embeddings of any other labels.
For the visual context embedding head, the positive examples are the embeddings of all images and image patches that contain the point under consideration, while the negative examples are embeddings of images that do not contain that point.
Similar to CLIP~\cite{radford2021learning}, we also note that a larger batch size helps reduce the variance in the contrastive loss function.
We use a batch size of $12,544$ everywhere since that is the maximum batch size we could fit in our VRAM of an NVIDIA Quadro GP100 GPU.

\section{Experimental Evaluation}
We evaluate \MODEL{} in terms of instance and semantic segmentation in images first -- to show that given ground truth data, it can learn meaningful scene representations. 
Then, we show that, only using weak web-model supervision, \MODEL{} can be used as a robot's spatial memory with semantic information.
Our visual segmentation experiments are performed on a subset of Habitat-Matterport 3D Semantic (HM3D semantics)~\cite{hm3d-semantics} dataset, while our robot experiments were performed on a Hello Robot Stretch using Hector SLAM~\cite{KohlbrecherMeyerStrykKlingaufFlexibleSlamSystem2011}.
We chose HM3D semantics as our sim testing ground because in this dataset, each scene comes with a different set of labels derived from free-form annotations.

\subsection{Instance and semantic segmentation in scene images}
The first task that we evaluate our model on is learning instance and semantic segmentation of 3D environments.
We assume that we have access to a scene, a collection of RGB-D images in it from different viewpoints, and a limited number of them are annotated either by humans, or by a model.
We consider two cases in this scenario: one where there are some human annotation data available, and in another where we are completely reliant on large, web-image trained models.

\subsubsection*{Baselines}
In our semantic and instance segmentation tasks, we use 2D RGB based segmentation models as our baselines.
In all of the few-shot segmentation experiments, we take a Mask-RCNN model with a ResNet50 FPN backbone, and a DeepLabV3 model with a ResNet50 backbone.
All baseline models were pre-trained on ImageNet-1K and then the COCO dataset.
We fine-tune the final layers of these pretrained models on each of our limited datasets, and then evaluate them on the held-out set.
For the RN50 FPN model, we report the mAP at [0.5-0.95] IoU range.
Detic is absent from the first two evaluations since it is a detection model and thus cannot be fine-tuned on segmentation labels.

\subsubsection*{Evaluating \MODEL{}}
Since \MODEL{} defines a function that maps from 3D coordinates, rather than from pixels, to representation vectors, to evaluate this model's learned representations we also have to use the depth and odometry information associated with the image.
To get semantic or instance segmentation, we take the depth image, using the camera matrix and odometry project it back to world coordinates, and then query the associated points in world coordinate from \MODEL{} to retrieve the associated representations with the points.
These representations can once again be projected back into the camera frame to reconstruct the segmentation map predicted by \MODEL{}.
Back-projecting to 3D world coordinates also lets \MODEL{} correctly identify visually occluded and obstructed instances in images, which is not easy for RGB-only models.

\begin{figure}[tb]
    \vskip -0.35cm
    \centering
    \includegraphics[width=\linewidth]{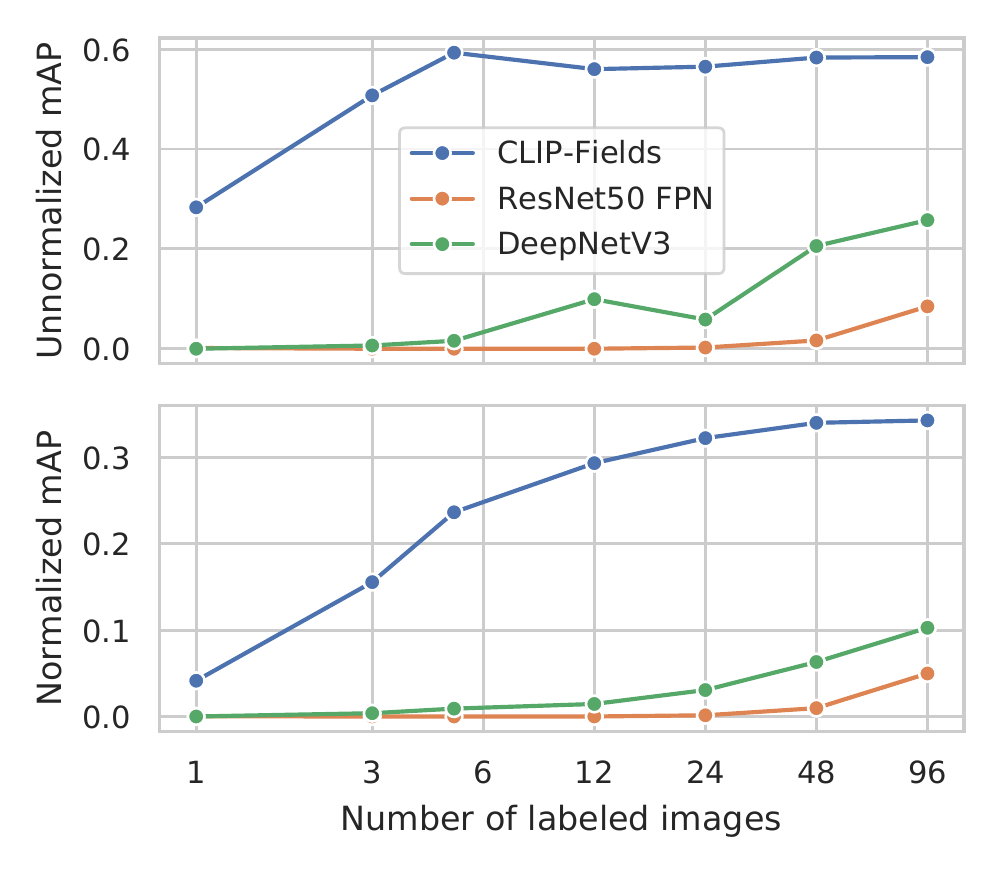}
    \caption{Mean average precision in instance segmentation on the Habitat-Matterport 3D (HM3D) Semantic dataset, (top) calculated over only seen instances, and (bottom) calculated over all instances.}
    \label{fig:inst_seg}
    \vskip -0.45cm
\end{figure}

\subsubsection{Low-shot instance identification}
In this setting, we assume that we have access to a few images densely annotated with an instance segmentation with associated instance IDs.
Such annotations are difficult for a human to provide, and thus it is crucial in this setting to perform well with very few (1-5) examples.

On this setting, we train \MODEL{} with the provided instance segmented RGB-D images and the associated odometry data, and compare with the baseline pretrained 2D RGB models fine-tuned on the same data.

As we can see in Figure~\ref{fig:inst_seg}, the average precision of the predictions retrieved from \MODEL{} largely outperforms the RGB-models.
This statement holds true whether we normalize by the number of seen instances in the training set or by the total number of instances in the scene.

\subsubsection{Low-shot semantic segmentation} 
\begin{figure}[tb]
    \centering
    \includegraphics[width=\linewidth]{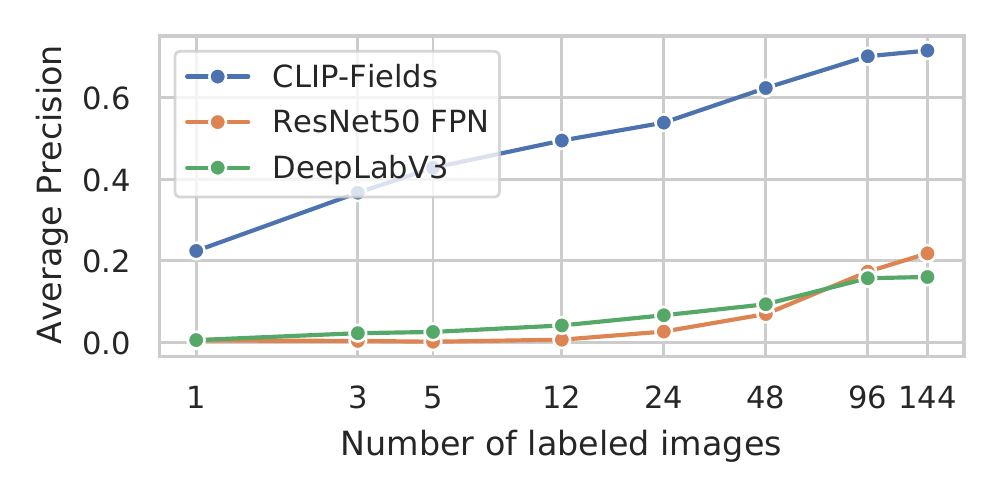}
    \vskip -0.1cm
    \caption{Mean average precision in semantic segmentation on the Habitat-Matterport 3D (HM3D) Semantic dataset. Here, the average precision numbers are averaged over all semantic classes.}
    \label{fig:gt_sem_seg}
    \vskip -0.4cm
\end{figure}
Next, we focus on a similar setting on semantically segmenting the views from the scene from a few annotations.

In Figure~\ref{fig:gt_sem_seg}, we see once again that \MODEL{} outperforms the RGB-based models significantly, to the point where even with three labelled views, \MODEL{} has a higher AP than any of the baseline RGB models.

\subsubsection{Zero-shot semantic segmentation}
\begin{figure}[tb]
    \centering
    \includegraphics[width=\linewidth]{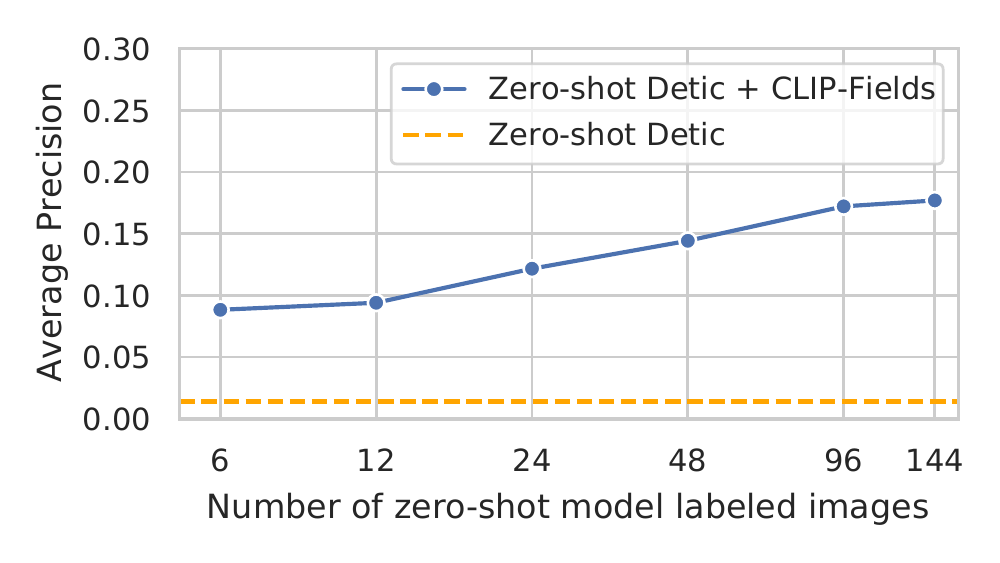}
    \caption{Mean average precision in zero-shot semantic segmentation on the Habitat-Matterport 3D (HM3D) Semantic dataset.}
    \label{fig:zero_shot_semseg}
    \vskip -0.2cm
\end{figure}

To examine the benefits derived purely from imposing multi-view consistency and a 3D structure over 2D model predictions, we experiment with \MODEL{} trained solely with labels from large web-image trained models in a zero-shot settings.
In this experiment, we train \MODEL{}  only with labels given to us by such large web models, namely Detic~\cite{zhou2022detecting}.
We get the labels by using Detic on the unlabeled training images, and then train \MODEL{} on it.
Besides text labels from Detic, we also use the CLIP visual representations to augment the implicit model, as described in Section~\ref{sec:objectives}.

As a baseline, we compare the trained \MODEL{} with performance of the same Detic model used to label the scene images.
Both \MODEL{} and the baseline had access to the list of semantic labels in each scene with no extra annotations.
We see in Figure~\ref{fig:zero_shot_semseg} that enforcing 3D structure and multi-view consistency in our segmentation predictions improves the test-time predictions considerably.

In all our visual segmentation experiments, we see that enforcing 3D consistency and structure using \MODEL{} helps identifying scene properties from images.
Back-projecting the rays can also help \MODEL{} correctly identify objects which are occluded and partially visible.
This property can be extremely helpful in a busy indoor setting where not every object can be visible from every angle.
Ability to work with occluded views and partial information can be a strong advantage for any embodied intelligent agent. 

\subsubsection{\MODEL{}'s robustness to label errors}
\label{sec:robustness}

\begin{figure}[tb]
    \centering
    \includegraphics[width=\linewidth]{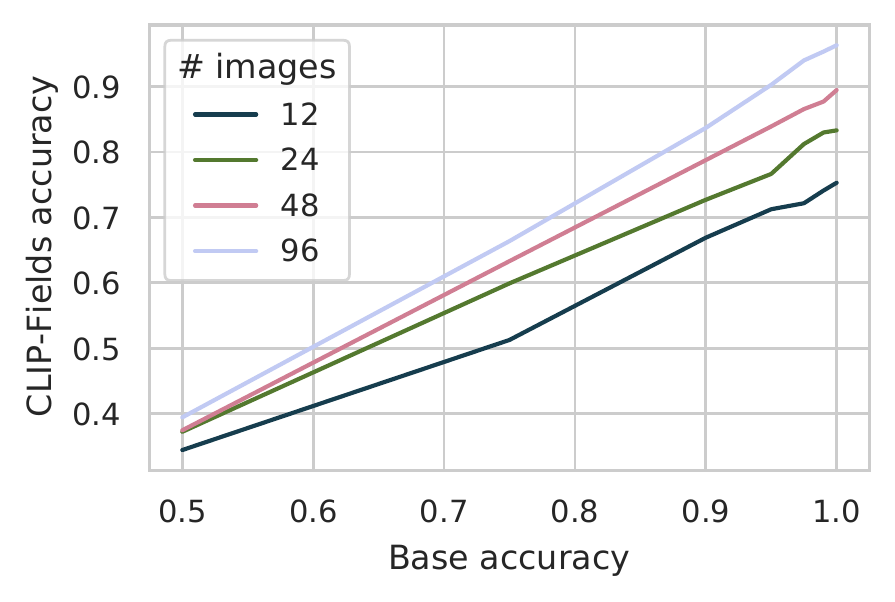}
    \caption{Mean average accuracy on the semantic segmentation task on the HM3D Semantic dataset with label noise simulating errors in base labelling models. Different lines show performance of models trained with a different number of labeled training frames.}
    \label{fig:robustness}
    \vskip -0.4cm
\end{figure}

In real-world applications, \MODEL{} relies on labels given by large-scale web-data trained models, which rarely (if ever) have perfect accuracy.
In this section, we examine the robustness of \MODEL{} to such label errors.
In this experiment, we simulate label errors by taking ground truth semantic labels in simulation, and for each frame and each object in that frame, flipping that object's label to another random label with probability $p$. 
By doing so, we simulate labelling our training data by a model whose mean accuracy is $1 - p$.

We see from Figure~\ref{fig:robustness} that as the base model's semantic label prediction accuracy increases, \MODEL{}'s label prediction accuracy increases almost linearly.
Importantly, there is no dramatic accuracy decrease when the base model accuracy goes below $1$.
Thus, we can see that \MODEL{} maintain reasonable accuracy as long as the base models are also reasonably accurate, which is the case for the state-of-the-art detection and segmentation models.
As the base models naturally improve over time with continuous efforts in the computer vision and natural language processing fields, we expect \MODEL{}'s performance to improve correspondingly.

\subsubsection{View Localization}

\begin{figure}[tb]
    \centering
    \includegraphics[width=\linewidth]{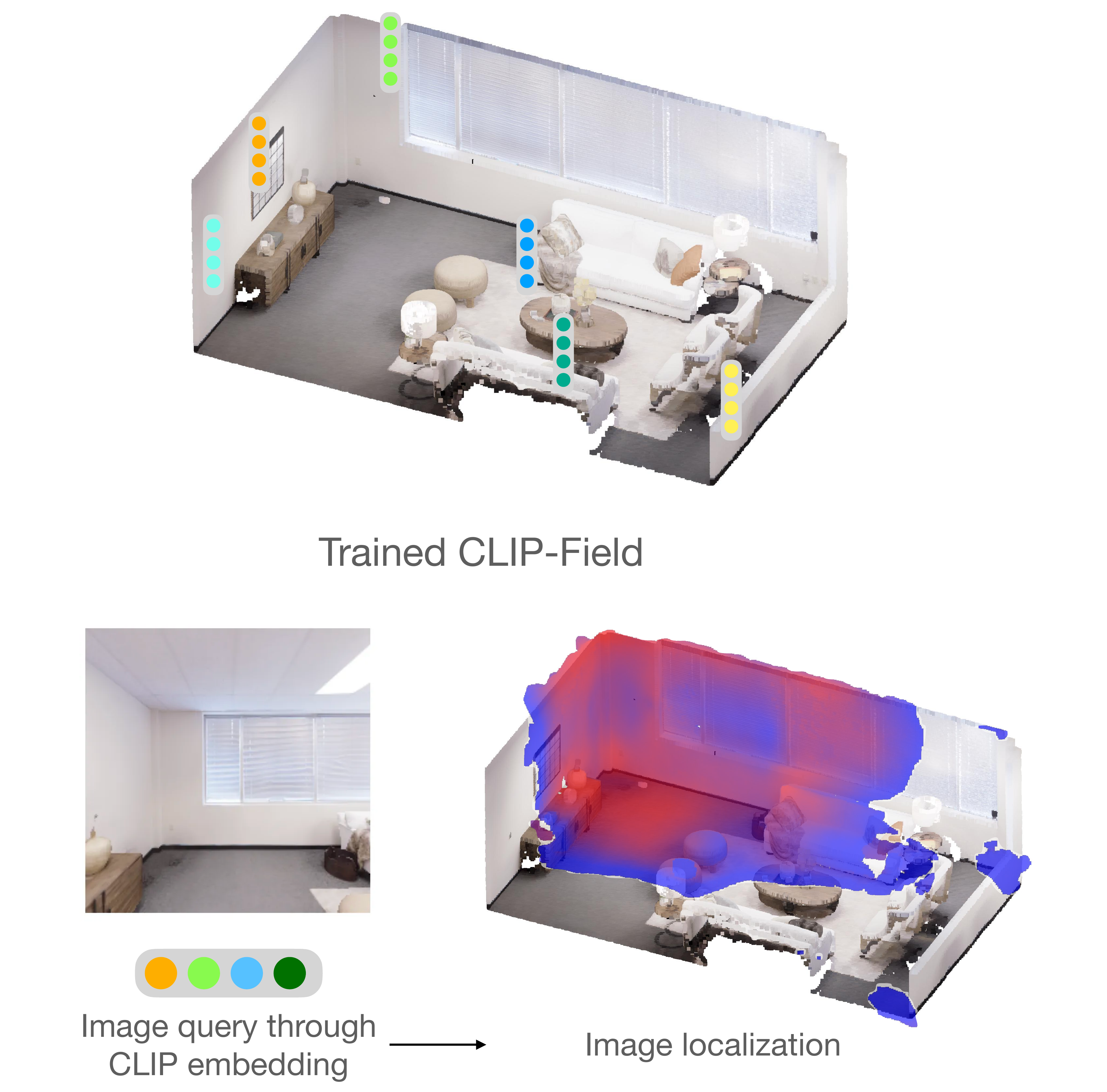}
    \caption{View localization using a trained \MODEL{}. We encode the query image on the bottom left to its CLIP representation, and visualize the locations whose \MODEL{} representations have the highest (more red) dot product with the embedded image. Lower dot products are blue; and below a threshold are uncolored.}
    \label{fig:snef_image_query}
    \vskip -0.75cm
\end{figure}

Since \MODEL{} is trained with CLIP embeddings at each coordinate, we can use such embeddings to localize an arbitrary view from the scene.
To do so, we simply find the CLIP embedding of the query image.
Then, we query the visual representation of the points in the scene, and take the dot product between the query representation and the point representations.
Due to the contrastive loss that CLIP was trained with, points that have similar embeddings to the query embedding will have the highest dot product.
We can use this principle to localize any view in the scene, as shown in Figure~\ref{fig:snef_image_query}.

\subsection{Semantic Navigation on Robot with \MODEL{} as Semantic-Spatial Memory}
\label{sec:robot-exp}

\begin{figure}[tb]
    \centering
    \includegraphics[width=\linewidth]{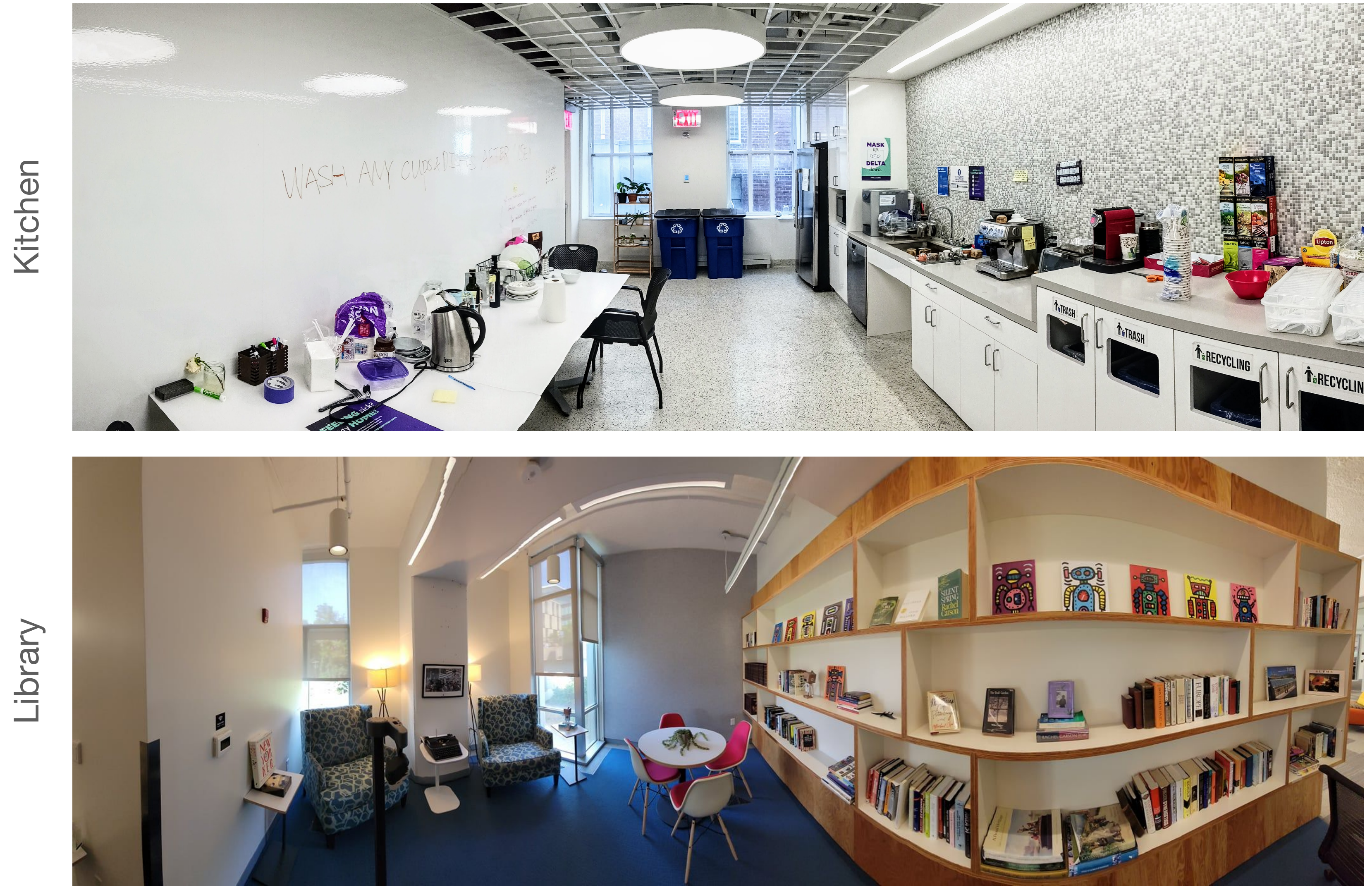}
    \caption{Scenes for our real-world semantic navigation experiments. The top scene is a lab kitchen and the bottom is a library/lounge.}
    \label{fig:experiment_settings}
    \vskip -0.5cm
\end{figure}

Training a \MODEL{} with available data, whether they are labeled by humans or pretrained models, gives us a mapping from real world coordinates to a vector representation trained to contain their semantic and visual properties (Section~\ref{sec:objectives}).
In this section, we evaluate the quality of the learned representations by using the learned model for downstream robot semantic navigation tasks.

\subsubsection{Task setup} We define our robot task in a 3D environment as a ``Go and look at \textit{X}'' task, where \textit{X} is a natural language query defined by the user.
To test \MODEL{}'s semantic understanding capabilities, we formulate the queries from three different categories:
\begin{itemize}
    \item \textit{Literal queries:} At this level, we choose \textit{X} to be the literal and unambiguous name of an object present in the scene, such as ``the refrigerator'' or ``the typewriter''.
    \item \textit{Visual queries:} At this level, we add references to objects by their visual properties, such as ``the red fruit bowl'' or ``the blue book with a house on the cover''.
    \item \textit{Semantic queries:} At this level, we add references to objects by their semantic properties, such as ``warm my lunch'' (microwave), or ``something to read''  (a book).
\end{itemize}

\begin{figure*}[!htb]
    \includegraphics[width=\textwidth]{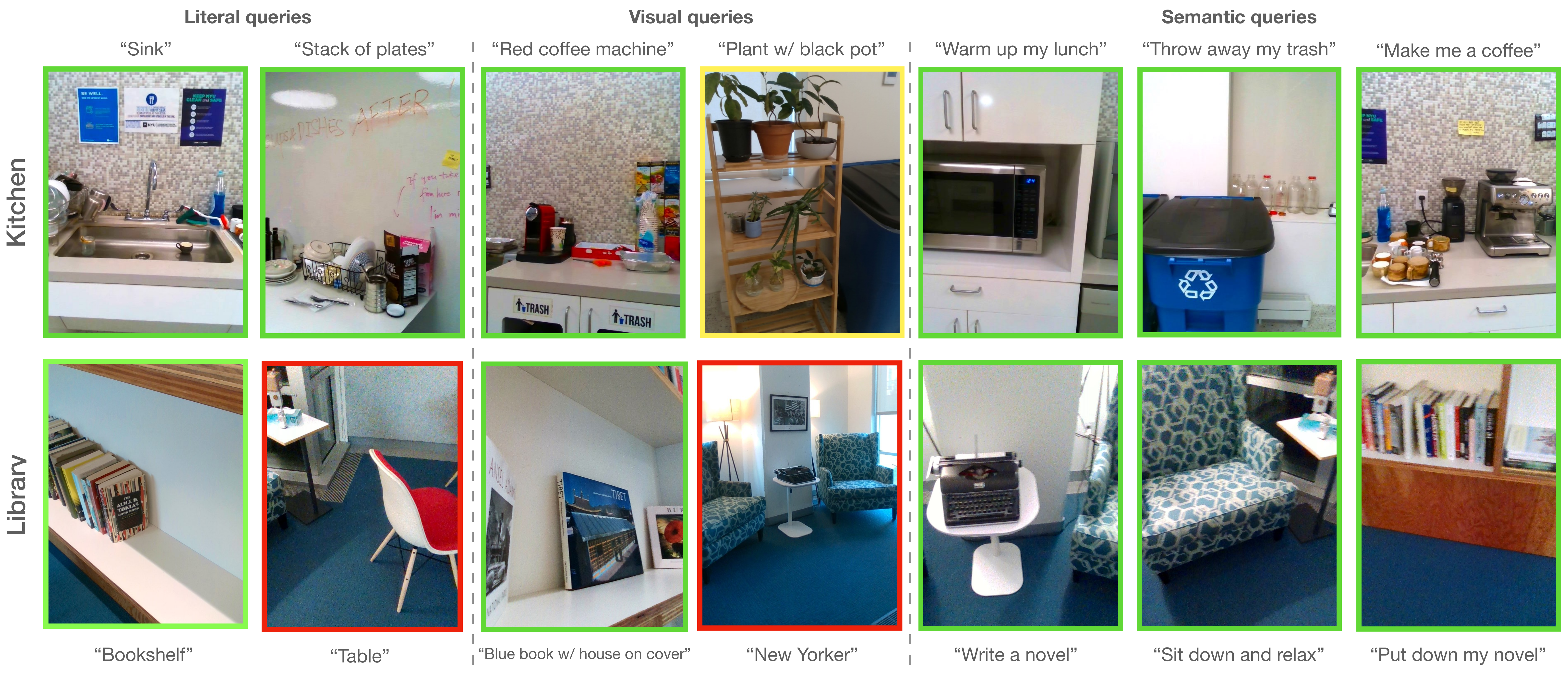}
    \caption{Examples of the robot's semantic navigation in two different testing environments, looking at objects given different queries. The images show the robot's POV given the associated query, with color coded borders showing approximate correctness. The rows show different two different scenes, top being in a lab kitchen and the bottom in our lab's library/lounge space, shown in detail in figure~\ref{fig:experiment_settings}.}
    \label{fig:query_results}
\vskip -0.25cm
\end{figure*}

\subsubsection{Data collection and training}
We ran our robot experiment in two different scenes, one in the lab kitchen, and another in the lab library (Figure~\ref{fig:experiment_settings}). 
For each of the scenes, we collected the RGB-D and odometry data with an iPhone 13 Pro with LiDAR sensors. 
The iPhone recording gave us a sequence of RGB-D images as well as the approximate camera poses in real world coordinate.  
On each of these scenes, we labelled a subset of the collected RGB images with Detic~\cite{zhou2022detecting} model using ScanNet200~\cite{rozenberszki2022language} labels.
Then, we created a training dataset with 3D world coordinates and their associated semantic and visual embeddings using the method described in Section~\ref{sec:dataset}.
On this dataset, we trained a \MODEL{} to synthesize all the views and their associated labels.

\begin{figure}[tb]
    \centering
    \includegraphics[width=\linewidth]{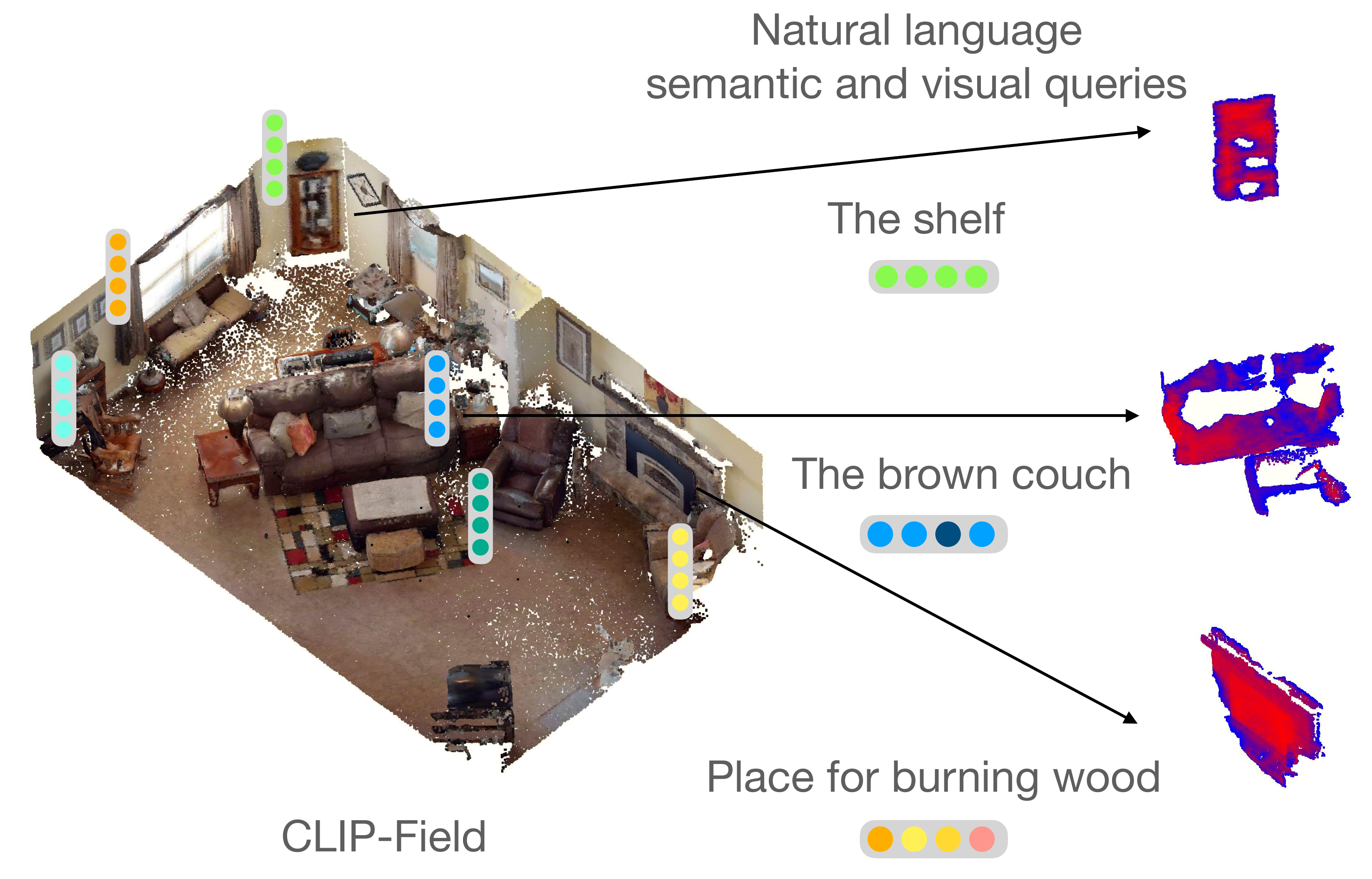}
    \caption{Running semantic queries against a trained \MODEL{}. We encode our queries with language encoders, and compare the encoded representation with the stored representation in \MODEL{} to then extract the best matches.}
    \label{fig:snef_query}
    \vskip -0.5cm
\end{figure}

\subsubsection{Robot execution}
Next, on our robot, we load the CLIP-Field to help with the localization and navigation of the robot.
When the robot gets a new text query, we first convert it to a representation vector.
We use Sentence-BERT to retrieve the semantic part of the query representation and CLIP text model to retrieve the vision-aligned part of the query representation.
Then, we find the coordinates of the point $P$ in space that has the highest alignment with the query representations, as described in Section~\ref{sec:problem_statement} and Figure~\ref{fig:snef_query}.
We use the robot's Hector SLAM~\cite{KohlbrecherMeyerStrykKlingaufFlexibleSlamSystem2011} navigation stack to navigate to that region, and point the robot camera to an XYZ coordinate where the dot product was highest.
We consider the navigation task successful if the robot can navigate to and point the camera at an object that satisfies the query.
We run twenty queries in the kitchen and fifteen queries in the library environment.

\subsubsection{Experiment results} 

In our experiments (Figure~\ref{fig:query_results}), we see that \MODEL{} let the robot navigate to different points in the environment from semantic natural language queries.
We generally observe that if an object was correctly identified by the web-image models during data preparation, when queried literally \MODEL{} can easily understand and navigate to it, even with intentional misspellings in the query.
However, if an object was misidentified during data preparation, \MODEL{} fails to correctly identify it as well.
For example, in row two, column two of Figure~\ref{fig:query_results}, the part of the floor that is identified as a ``table" was identified as a ``table" by our web-image model earlier.
This observation lines up with our simulated experiments in Section~\ref{sec:robustness} where we saw that \MODEL{} performance has a linear relationship with the base models' performance.
For semantic queries, \MODEL{} sometimes confuses two related concepts; for example, it retrieves the dishwasher for both ``place to wash my hand" and ``place to wash my dishes". %
Finally, the visual queries sometimes put a higher weight on the semantic match rather than visual match, such as retrieving a white fruit bowl for "red fruit bowl" instead of the red bowl in the scene. However, the right object is retrieved if we query for "red plastic bowl".

We have presented detailed logs of running \MODEL{} on the robot in the kitchen environment in Appendix~\ref{sec:app:exp_logs} detailing all the queries and the resulting robot behavior.

\section{Conclusions and Future Work}

We showed that \MODEL{} can learn 3D semantic scene representations 
from little or no labeled data, relying on weakly-supervised web-data trained models, and that we can use these model to perform a simple ``look-at'' task in the real world.
\MODEL{} allow us to answer queries of varying levels of complexity. %
We expect this kind of 3D representation to be generally useful for robotics.
For example, it may be enriched with affordances for planning; the geometric database can be %
combined with end-to-end differentiable planners.
In future work, we hope to explore models that share parameters across scenes, and can handle dynamic scenes and objects.

\bibliographystyle{plainnat}
\bibliography{references}

\clearpage
\section{Appendix}

\subsection{Training details}
We release our open source code at the Github repo \url{https://github.com/clip-fields/clip-fields.github.io} with full details about how to train a new \MODEL{} on any environment.
The code is also shared in the attached supplementary information zip file.
While the published code should be sufficient to reproduce our work and experiments, we are describing the most important training details and hyperparameters here for reproducibility purposes.

\begin{table}[hbt]
\caption{Optimization hyperparameters}
\centering
\begin{tabular}{@{}lc@{}}
\toprule
Parameter & Value \\ \midrule
Optimizer          & Adam      \\
Learning rate          & $10^{-4}$      \\
Weight decay          & $3 \times 10^{-3}$      \\
$\beta$          & $(0.9, 0.999)$      \\
Learning rate schedule         & None      \\
Epochs          & $100$      \\
Per epoch iters          & $3 \times 10^6$      \\
Batch size         & $12,544$      \\
$\alpha$ (Sec.~\ref{sec:objectives}, when applicable)          & $100.0$      \\
\end{tabular}
\end{table}

\begin{table}[hbt]
\caption{Architecture and Instant-NGP hyperparameters}
\centering
\begin{tabular}{@{}lc@{}}
\toprule
Parameter & Value \\ \midrule
Intermediate representation dimension          & $144$      \\
NGP grid levels          & $18$      \\
NGP per-level scale          & $2$       \\
NGP level dimension          & $8$      \\
NGP $\log_2$ hash map size          & $20$      \\
MLP number of hidden layers          & $1$      \\
MLP hidden layer size          & $600$      \\
\end{tabular}
\end{table}

\begin{table}[bt]
\caption{External model configurations}
\centering
\begin{tabular}{@{}lccl@{}}
\toprule
Task & Model & Instance \\ \midrule
Object detector          & Detic      & CLIP + SwinB      \\
Vision-language model          & CLIP      & ViT-B/32     \\
Language model          & Sentence-BERT    & all-mpnet-base-v2     \\
\end{tabular}
\end{table}

\subsection{Real world experiment logs}
\label{sec:app:exp_logs}
In this section, we reproduce the exact real-world qualitative observations that we made by running our robot on the Kitchen scenario.
We present this for the readers to get a full picture of what the robot queries looked like, and how the \MODEL{} responded to each of the queries.

\begin{enumerate}
    \item Literal queries:
    \begin{enumerate}
        \item Stack of plates: success, found the dishwashing rack with plates in it.
        \item Microwave: success, found the microwave oven in the lab kitchen.
        \item The fridghe (misspelling intentional): success, found the large standing fridge in the corner.
        \item Coffee machine (ambiguous query): success, found the silver coffee maker.
        \item Sink: success, found the sink.
        \item Toaster oven: failure, found the microwave oven instead of the toaster oven.
    \end{enumerate}
    \item Visual queries:
    \begin{enumerate}
        \item White ceramic bowl: success, found the bowl by the dishwashing rack.
        \item Red plastic bowl: success, found the red bowl above the trash cabinets.
        \item Red fruit bowl: failure, found the white bowl by the dishwashing rack.
        \item Espresso machine: success, found the nespresso machine by the coffee machine.
        \item Blue gargabe bin: success, found one of the two blue recycling bins in the kitchen.
        \item Potted plant in a black pot: success, ambiguous, found the potted plants in a shelf. Isolating the black flower pot was ambiguous since the robot doesn't get too close to scene objects.
        \item Purple poster: success, found the poster above the sink.
    \end{enumerate} 
    \item Semantic queries: 
    \begin{enumerate}
        \item Wash my dishes: success, finds the dishwasher as intended.
        \item Wash my hand: failure, finds the dishwasher instead of the sink.
        \item Throw my trash: success, finds the recycling bins (although not entirely climate friendly behavior.)
        \item Put away my leftovers: failure, pointed the camera at the trash cabinet instead of the fridge or the cabinets. Potentially because the trash cabinets got identified as ``cabinets" by our detectors.
        \item Fill out my water bottle: success, finds the glass bottles at the corner of the kitchen. While the original intention was to find the water cooler, the response is reasonable.
        \item Make some coffee: success, found the coffee maker and grinders.
        \item Warm up my lunch: success, found the microwave oven.
    \end{enumerate}
\end{enumerate}

\end{document}